\documentclass[sigconf]{acmart}
\usepackage{hyperref}
\usepackage[linesnumbered,ruled]{algorithm2e}
\usepackage{multirow}

\AtBeginDocument{%
  \providecommand\BibTeX{{%
    \normalfont B\kern-0.5em{\scshape i\kern-0.25em b}\kern-0.8em\TeX}}}

\setcopyright{acmcopyright}
\copyrightyear{2020}
\acmYear{2020}
\acmDOI{10.1145/1122445.1122456}

\acmConference[AdKDD '20]{AdKDD '20}{August 23, 2020}{San Diego, CA}
\acmBooktitle{AdKDD '20, August 23, 2020, San Diego, CA}
\acmPrice{15.00}
\acmISBN{978-1-4503-XXXX-X/18/06}

\begin{document}

\title{Adversarial Validation Approach to Concept Drift Problem in User Targeting Automation Systems at Uber}

\author{Jing Pan}
\affiliation{%
  \institution{Uber Technologies}
  \city{San Francisco}
  \state{California}
  \postcode{94105}
}
\email{jing.pan@uber.com}

\author{Vincent Pham}
\affiliation{%
  \institution{Uber Technologies}
  \city{San Francisco}
  \state{California}
  \postcode{94105}
}
\email{vincent.pham@uber.com}

\author{Mohan Dorairaj}
\affiliation{%
  \institution{Uber Technologies}
  \city{San Francisco}
  \state{California}
  \postcode{94105}
}
\email{mohan@uber.com}

\author{Huigang Chen}
\affiliation{%
  \institution{Uber Technologies}
  \city{San Francisco}
  \state{California}
  \postcode{94105}
}
\email{huigang@uber.com}

\author{Jeong-Yoon Lee}
\affiliation{%
  \institution{Uber Technologies}
  \city{San Francisco}
  \state{California}
  \postcode{94105}
}
\email{jeong@uber.com}

\renewcommand{\shortauthors}{Pan, et al.}

\begin{abstract}
    In user targeting automation systems, concept drift in input data is one of the main challenges. It deteriorates model performance on new data over time. Previous research on concept drift mostly proposed model retraining after observing performance decreases. However, this approach is suboptimal because the system fixes the problem only after suffering from poor performance on new data. Here, we introduce an adversarial validation approach to concept drift problems in user targeting automation systems. With our approach, the system detects concept drift in new data before making inference, trains a model, and produces predictions adapted to the new data. We show that our approach addresses concept drift effectively with the AutoML3 Lifelong Machine Learning challenge data as well as in Uber's internal user targeting automation system, MaLTA.
\end{abstract}



\maketitle

\section{Introduction}
At Uber, user targeting automation system - MaLTA (Machine Learning based Targeting Automation) is responsible for building and deploying targeting models for internal stakeholders. These models are used to identify and target existing users with the most relevant communication, promotion, or marketing advertisement for our different lines of business. These models often have different predicting goals and use multiple data sources including both raw and derived features. Furthermore, they usually have different life cycles. A model could be built, deployed, used, deprecated, and resurrected at different points in time. Such modeling aspects are of the utmost importance to the accuracy and robustness of our system. In practice, however, they are often outside the control of the user targeting automation system due to business and organizational challenges. The predictions from our models are commonly used in other systems many weeks out in the future without any immediate observable feedback. 

\begin{figure}[tbp]
    \centering
    \includegraphics[width=0.5\textwidth]{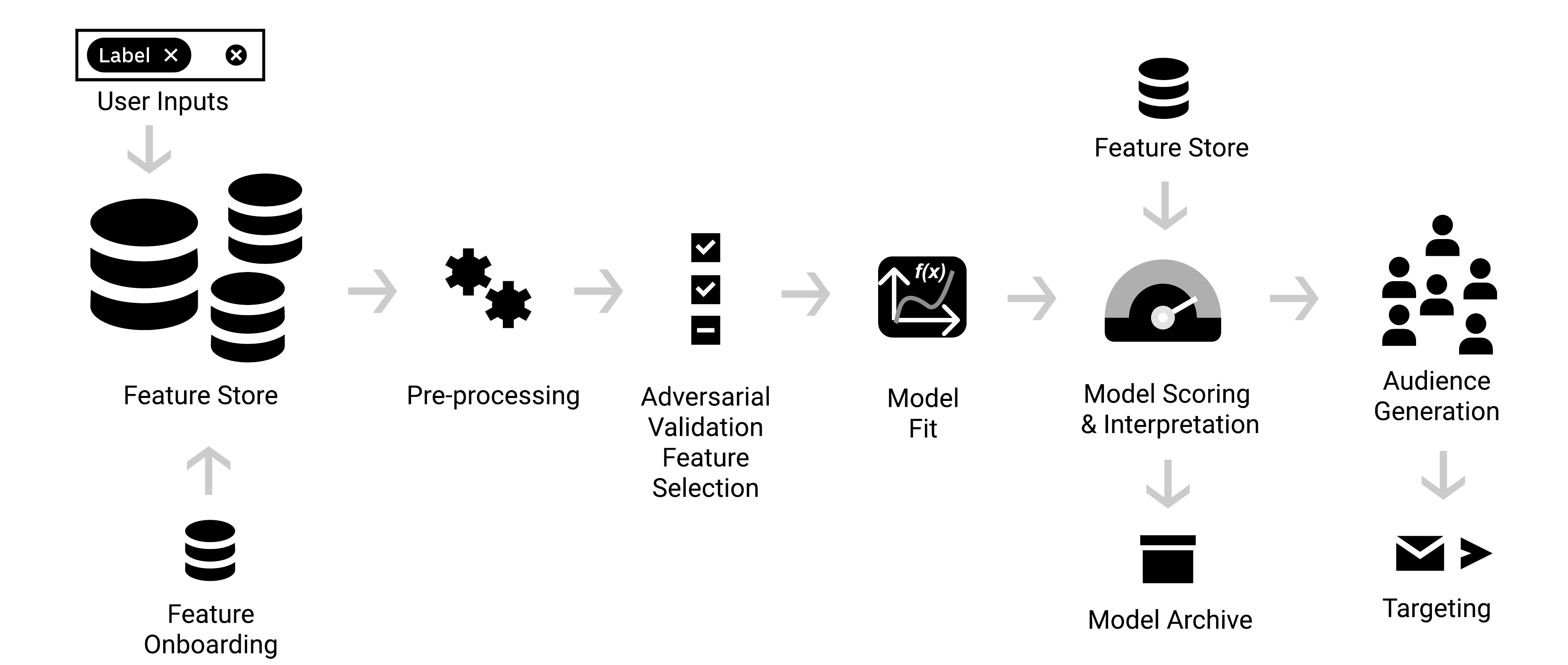}
    \caption{MaLTA model training flow with adversarial validation based on feature selection}
    \label{figure:fig2}
\end{figure}

In MaLTA, we leverage multiple snapshots of the user's past behavior, normally expressed through handcrafted derived features and metrics produced by different business units, to model the evolution of the user's future behaviors. This process, however, can cause concept drift as the distribution of these derived features are could shift over times. This issue is inherent to the data collection and modeling process. To overcome this problem, we use adversarial validation in the deployed system (Fig~\ref{figure:fig2}).

\begin{figure}[tbp]
    \centering
	\includegraphics[width=0.5\textwidth]{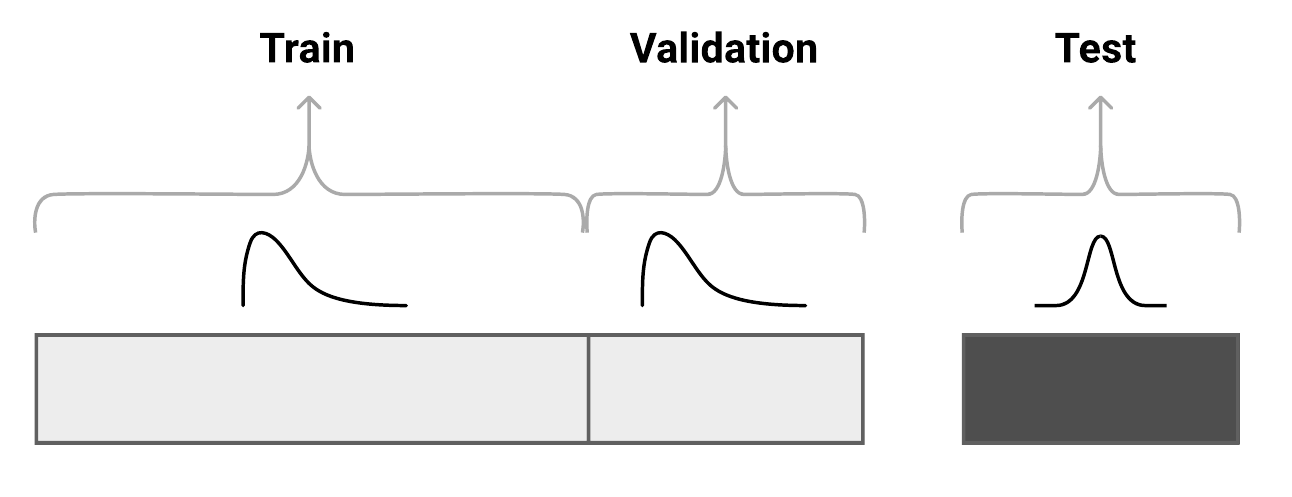}
	\caption{An example of different feature distribution between the train and test data}
    \label{figure:fig1}
\end{figure}

Adversarial validation is an approach to detect and address the difference between the training and test datasets. During machine learning model development, the model performance on part of the training dataset, the validation dataset, is used as a proxy of the performance on test data. However, if the feature distributions in the training and test datasets are different (e.,g. see Fig~\ref{figure:fig1}), the performance on the validation and test datasets will be different. In adversarial validation, a binary classifier, adversarial classifier, is trained to predict if a sample belongs to the test dataset. Classification performance better than random guess indicates the different feature distributions between the training and test datasets. Furthermore, the adversarial classifier can be used to balance the training and test datasets and improve the model performance on the test dataset.

Despite its practical importance and popularity in machine learning competitions \cite{Zajac2016}, to our best knowledge, no previous study has been published on adversarial validation. Here, we introduce adversarial validation and propose three adversarial validation methods to address the concept drift problem.

The paper is organized as follows. Section 2 discusses related concepts and efforts to address the concept drift issue. Section 3 describes three adversarial validation methods in detail. Experimental settings that use public data and Uber's internal data are presented in Section 4. Results that compare the performances of different adversarial validation methods are presented and discussed in Section 5. Section 6 concludes the paper with future directions. 

\section{Related Work}
\subsection{Generative Adversarial Networks}
The name of adversarial validation came from Generative Adversarial Networks (GAN) \cite{Goodfellow2014}, which became increasingly popular in content generation. It consists of two neural networks, the generator and discriminator. The generator produces new data resembling the training data provided, while the discriminator distinguishes between the original training data and newly generated data. 

Similarly, adversarial validation consists of two models. One model is trained to predict a target variable with training data, while the other model, an adversarial classifier is trained to distinguish between the training and test datasets. In adversarial validation, the predictions of the adversarial classifier are used to help the first model generalize better with the test data instead of generating the new test data as in GAN.

\subsection{Heterogeneous Treatment Effect Estimation}
Adversarial validation approach is similar to propensity score modeling in causal inference \cite{ROSENBAUM1983, Austin2011}. In causal inference, propensity score modeling addresses the heterogeneity between the treatment and control group data by training a classifier to predict if a sample belongs to a treatment group. Rosenbaum and Rubin argue in \cite{ROSENBAUM1983} that it is sufficient to achieve the balance in the distributions between the treatment and control groups by matching on the single dimensional propensity score alone, which is significantly more efficient than matching on the joint distribution of all confounding variables.

More generally, analogous to the model prediction problem with adversarial validation, heterogeneous treatment effect estimation aims to estimate the effect of a treatment variable on an outcome variable from observational data at the individual sample level. With randomized controlled trial (RCT) data, we can estimate the average treatment effect (ATE) by calculating the difference between the average outcomes of the treatment and control groups, since the control group forms a valid counterfactual prediction given the similar distribution to the treatment group. However, with observational data, the split between the treatment and control groups is not random and might depend on other confounding variables. It is generally known that a naive counterfactual prediction, such as T learner in \cite{Kunzel2019}, which uses a model whose training data is from one group with different distribution than the other group where the counterfactual prediction is made, generates bias. Therefore, to address this key issue, different methods have been designed to use the propensity score as a critical part of the estimation procedure  \cite{Kunzel2019, Shalit2017, Nie2017}. In adversarial validation, we also explore the use of the propensity score in the main prediction model to reduce the bias in the test dataset.

\subsection{AutoML3 for Lifelong Machine Learning Challenge}
At NeurIPS 2018, the AutoML3 for Lifelong Machine Learning challenge was hosted, where “the aim is assessing the robustness of methods to concept drift and its lifelong learning capabilities” \cite{Madrid2019}.

The winning solutions at AutoML3 used similar approaches to each other \cite{Wilson2020}. First, a model was trained only with the training data and predicted for test data. Then, after observing the model performance on the test data, models were retrained with new training data consisting of the old training and latest test data with labels using techniques such as incremental training and sliding window to weigh more on the latest data.

We argue that this is suboptimal because it addresses concept drift only after observing the model performance on the test data. With adversarial validation methods proposed in the next section, one can detect and address concept drift before making the inference on the test data.

\section{Adversarial Validation Methods}
Adversarial validation can be used to detect and address concept drift problem between the training and test data.

We start with a labeled training dataset $\{(y_{train}, X_{train})\} \in \mathbb{R} \times \mathbb{R}^d$, and an unlabeled test dataset $\{X_{test}\} \in \mathbb{R}^d$ with an unknown conditional probability $P_{y|X}$. Then, we train an adversarial classifier that predicts $P(\{train, test\}|X)$ to separate train and test, and generate the propensity score $p_{propensity} = P(test | X)$ on both $X_{test}$ and $X_{train}$.

The feature importance and propensity score from the adversarial classifier can be used to detect concept drift between the training and test data, and provide insights on the cause of the concept drift such as which features and subsamples in the training data are most different from ones in the test data.

In addition to concept drift detection, here, we propose three adversarial validation methods that address concept drift between the training and test data, and generate predictions adapted to the test dataset.

\subsection{Automated Feature Selection}
If distributions of the features from the train and test data are similar, we expect the adversarial classifier to be as good as random guesses. However, if the adversarial classifier can distinguish between training and test data well (i.e. AUC score $\gg$ 50\%), the top features from the adversarial classifier are potential candidates exhibiting concept drift between the train and test data. We can then exclude these features from model training.

Such feature selection can be automated by determining the number of features to exclude based on the performance of adversarial classifier (e.g. AUC score) and raw feature importance values (e.g. mean decrease impurity (MDI) in Decision Trees) as follows:

\begin{enumerate}
\item Train an adversarial classifier that predicts $P(\{train, test\}|X)$ to separate train and test.
\item If the AUC score of the adversarial classifier is greater than an AUC threshold $\theta_{auc}$, remove features ranked within top $x\%$ of remaining features in feature importance ranking and with raw feature importance values higher than a threshold $\theta_{imp}$ (e.g. MDI $>$ 0.1).
\item Go back to Step 1, if AUC score greater than $\theta_{auc}$.
\item Once the adversarial AUC drops lower than $\theta_{auc}$, train an outcome classifier with the selected features and original target variable.
\end{enumerate}

\begin{figure}[tbp]
    \centering
    \includegraphics[width=0.5\textwidth]{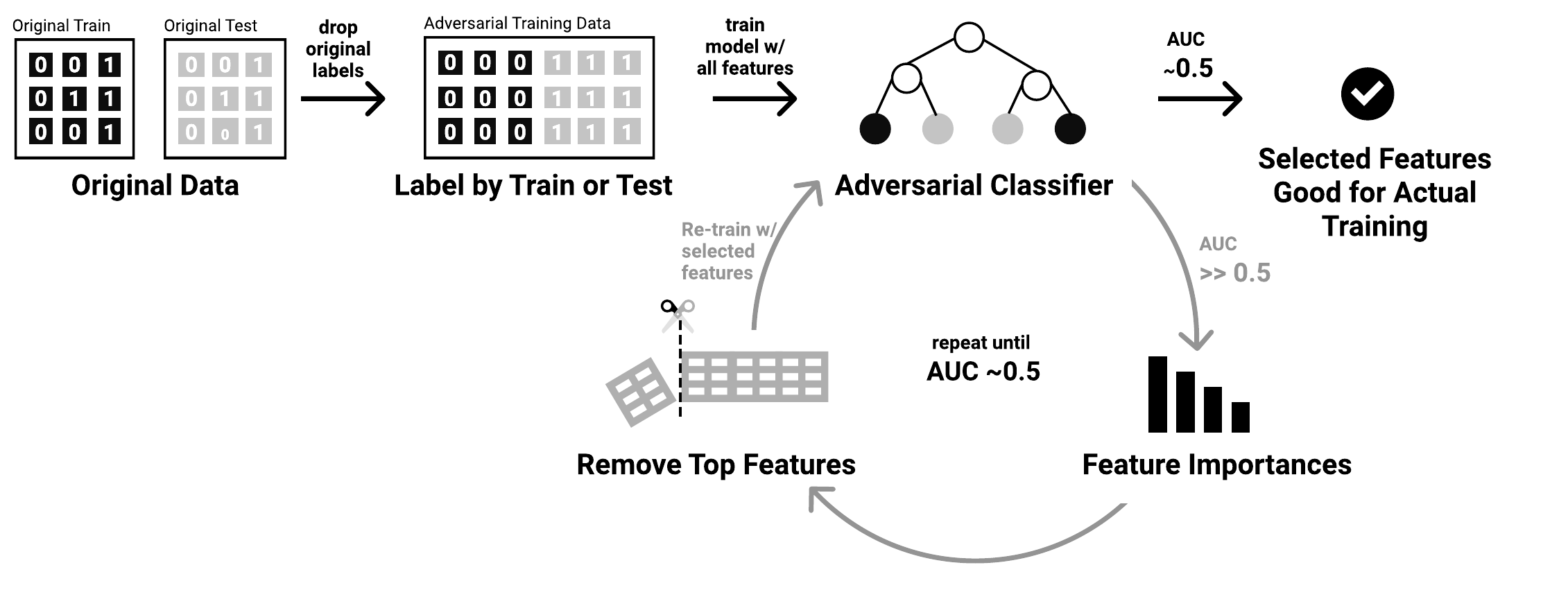}
    \caption{A diagram of adversarial validation with automated feature selection in MaLTA}
    \label{figure:fig3}
\end{figure}

Figure~\ref{figure:fig3} shows the diagram of adversarial validation with automated feature selection in MaLTA.

Automated feature selection prevents a model from overfitting to features with potential concept drift, and, as a result, leads to an outcome model that generalizes well on the test data. There is a trade-off for this method between losing information by dropping features from the model and reducing the size of training data using other approaches proposed below.\\

\subsection{Validation Data Selection}
With validation data selection, we construct a new validation dataset $\{(y_{val}, X_{val})\}$ by selecting from the training data so that the empirical distribution of the features data is similar to the test data, $P_{X_{val}} \approx P_{X_{test}}$. This way, model evaluation metrics on the validation set should get similar results on the test set, which means if the model works well on the validation data, it should work well on the test data.

Specifically, we apply propensity score matching (PSM) methodology \cite{Austin2011} to reduce the selection bias due to concept drift as follows:

\begin{enumerate}
\item Train an adversarial classifier that predicts $P(\{train, test\}|X)$ to separate train and test, and generate the propensity score $p_{propensity} = P(test | X)$ on both $X_{test}$ and $X_{train}$.
\item Run propensity score matching for the propensity scores with nearest neighbor and check the standardized mean difference (SMD) defined below for the propensity scores and covariates.
\begin{equation}
    SMD_x = \frac{E[x_{match}] - E[x_{test}]}{\sqrt{\frac{1}{2}(Var[x_{match}] + Var[x_{test}])}}.
\end{equation}
We consider the matched data to be balanced if $SMD < 0.1$. Tune the nearest neighbor threshold if necessary to achieve balance.
\item Pick the matched examples (e.g. 20\% of train) as the new validation dataset $\{(y_{val}, X_{val})\}$ and the remaining as the train data to train the model.
\end{enumerate}

It is possible that there are not enough matched samples from the training dataset because the training and test features are significantly different. In that case, we can select the subset of training data with the highest propensity scores $p_{propensity}$ as in \cite{Zajac2016}.

\subsection{Inverse Propensity Weighting}
Alternative to the validation data selection method that extracts a matched validation set while maintaining the training set the same, one can also use the inverse propensity weight (IPW) \cite{Austin2011} technique to generate the weights for the training set for weighted training. Specifically, we use the weights from the propensity score for the training sample as follows:

\begin{equation}
    w = \frac{p_{propensity}}{1 - p_{propensity}}.
\end{equation}

Therefore, the weighted distribution of the features in the training data follows

\begin{equation}
\begin{split}
    p(X|train) \times w & = p(X|train) \times \frac{P(test | X)}{P(train | X)} \\
    				& \propto P(test|X)p(X) \propto P(X|test),
\end{split}
\end{equation}
which reproduces the distribution of the features in the test data. We trim the weights for those observations with propensity scores close to 1 to avoid the pathological case of over-reliance of those observations and the consequential reduction of the effective sample size.

\begin{table}[tbp]
\centering
\caption{The parameters of algorithms used in experiments}
\label{table:tab0}
\resizebox{\columnwidth}{!}{
\begin{tabular}{c c p{30mm}}
\toprule
Experiment & Algorithm & Parameters \\
\midrule
\multirow{2}{*}{AutoML3} & GBDT (LightGBM) & max\_depth=5, \newline 
subsample=0.5, \newline 
subsample\_freq=1, \newline 
colsample\_bytree=0.8, \newline 
reg\_alpha=1, \newline 
reg\_lambda=1, \newline 
importance\_type='gain', \newline
early\_stopping\_rounds=10 \\
& RF \& DT (scikit-learn) & min\_samples\_leaf=20, min\_impurity\_decrease=0.01 \\
\midrule
\multirow{2}{*}{MaLTA} & GBDT (LightGBM) & num\_leaves=15, \newline 
reg\_alpha=1, \newline 
reg\_lambda=1, \newline 
importance\_type='gain', \newline
early\_stopping\_rounds=10 \\
& RF \& DT (scikit-learn) & min\_samples\_leaf=20, min\_impurity\_decrease=0.01 \\
\bottomrule
\end{tabular}}
\end{table}

\section{Experiments}
Adversarial validation with three different methods, feature selection, validation selection, and inverse propensity weighting (IPW) are applied to seven datasets from AutoML3 for Lifelong Machine Learning Challenge as well as MaLTA dataset. For adversarial validation with feature selection, three different algorithms, Decision Trees (DT) \cite{steinberg2009cart}, Random Forests (RF) \cite{breiman2001random}, and Gradient Boosted Decision Trees (GBDT) \cite{friedman2001greedy} are used for model training.

For outcome classifiers with the original target variables as target, we use \texttt{LightGBM} \cite{ke2017lightgbm} to train GBDT models. For adversarial classifiers with the train / test split as target, we use \texttt{LightGBM} for GBDT models, and \texttt{scikit-learn} \cite{pedregosa2011scikit} for DT and RF  models.

As feature engineering, missing values in numerical features are replaced with zero for \texttt{scikit-learn} models that require missing value imputation. Categorical features are label-encoded with missing values as a new label.

For both the outcome and adversarial classifiers, decision-tree-based algorithms are chosen to minimize feature engineering across datasets, which might affect the model performances. Parameters of algorithms are fixed in each experiment across the datasets and adversarial validation methods. Default parameters of GBDT (\texttt{LightGBM}), and RF and DT (\texttt{scikit-learn}) are used except for regularization parameters, which are set to avoid overfitting. Parameters different from default values are shown in Table~\ref{table:tab0}.

 GBDT models are trained with early stopping: 25\% of the training dataset is used as the validation set, and GBDT models train until the performance on the validation set stops improving. In the validation data selection and IPW methods, adversarial classifiers are trained with 5-fold cross-validation, and out-of-fold predictions are used as propensity scores.
 
\subsection{AutoML3 for Lifelong Machine Learning Challenge Datasets}

\begin{figure}[tbp]
    \centering
    \includegraphics[width=0.5\textwidth]{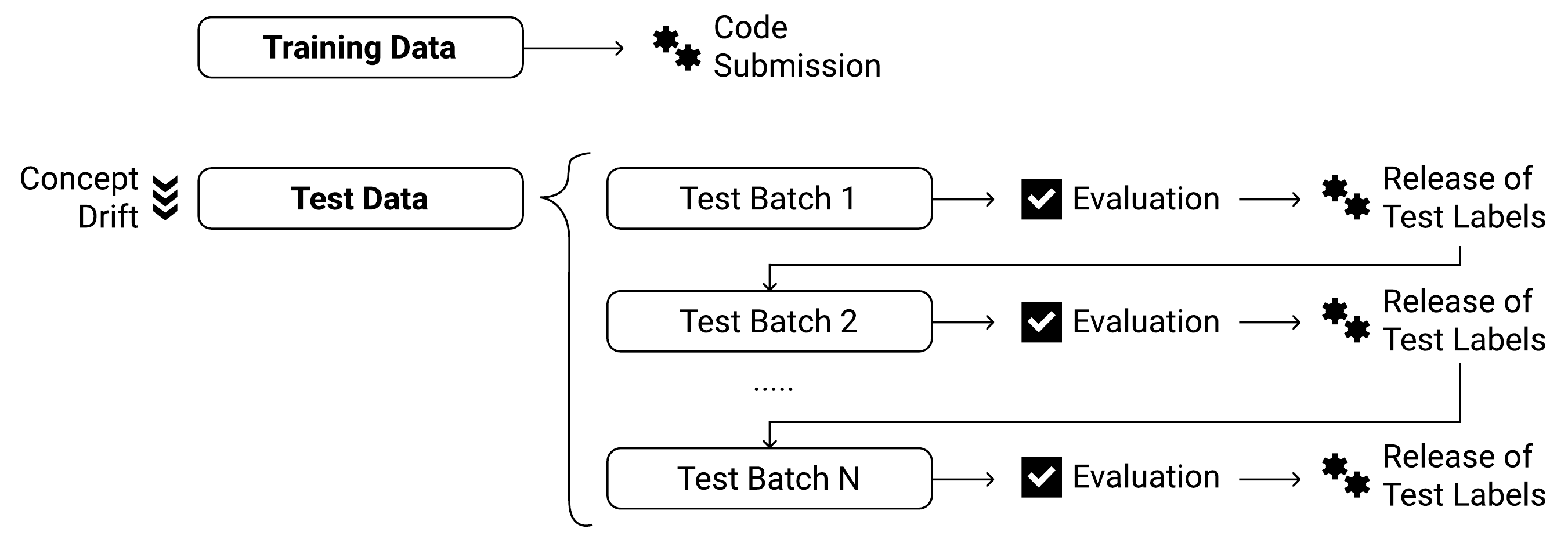}
    \caption{Evaluation scenario considered in the AutoML3 challenge \cite{Madrid2019}}
    \label{figure:fig4}
\end{figure}

At the AutoML3 challenge, participants started with a labeled training dataset, and a series of test datasets were provided sequentially. Once participants submitted predictions for a test dataset, the labels for the test dataset were released, and the next test dataset without label was available (see Figure~\ref{figure:fig4}). The model performance determined by taking the average performance across all test datasets.

We use public datasets made available during the feedback phase at AutoML3. The public datasets consist of seven datasets: \texttt{ADA}, \texttt{RL}, \texttt{AA}, \texttt{B}, \texttt{C}, \texttt{D} and \texttt{E}. Dataset \texttt{ADA} and \texttt{RL} have one training and three test datasets, while datasets \texttt{AA}, \texttt{B}, \texttt{C}, \texttt{D} and \texttt{E} have one training and four test datasets. All seven datasets have binary target variables.

The description of datasets, such as the number of features, size of training and test datasets, percentage of missing values after label encoding of categorical features, and adversarial validation AUC scores are shown in Table~\ref{table:tab1}. All datasets except \texttt{ADA} show perfect (100\%) or almost perfect adversarial validation AUC scores, indicating different feature distributions between the training and test datasets. On the other hand, \texttt{ADA} shows close-to-random (49\%) adversarial validation AUC score, indicating the consistent feature distribution between the training and test datasets.

For a demonstration purpose, we exclude date-time and multi-value categorical features, and use only numerical and categorical features for model training. The AutoML3 datasets are available at \url{https://competitions.codalab.org/competitions/19836}. The code for the experiment with the AutoML3 datasets is available at \url{http://t.uber.com/av-github}.

\begin{table}[tbp]
\centering
\caption{AutoML3 feedback phase public datasets}
\label{table:tab1}
\resizebox{\columnwidth}{!}{
\begin{tabular}{p{28mm} r r r r r r r}
\toprule
& ADA & RL & AA & B & C & D & E\\
\midrule
\# of Total Features & 48 & 22 & 82 & 25 & 79 & 76 & 34\\
Numerical Features & 48 & 14 & 23 & 7 & 20 & 55 & 6\\
Categorical Features & 0 & 8 & 51 & 17 & 44 & 17 & 25\\
Training Dataset Size & 4K & 31K & 10M & 1.6M & 1.8M & 1.5M & 16M \\
Test Dataset \#1 Size & 41K & 5K & 9M & 1.7M & 1.9M & 1.5M & 17M \\
Test Dataset \#2 Size & 41K & 5K & 9M & 1.6M & 1.3M & 1.6M & 18M\\
Test Dataset \#3 Size & 41K & 15K & 10M & 1.4M & 1.6M & 1.5M & 18M\\ 
Test Dataset \#4 Size & - & - & 9M & 1.7M & 1.8M & 1.6M & 18M \\
Missing Values \% & 0\% & 4\% & 0\% & 2\% & 3\% & 1\% & 0\% \\
Adversarial Validation \newline AUC Score & 49\% & 98\% & 100\% & 98\% & 100\% & 100\% & 100\%  \\
\bottomrule
\end{tabular}}
\end{table}

\subsection{MaLTA Datasets}
MaLTA constructs automated machine learning models to target users who show a higher propensity score toward being cross-sold into new products and services. For this experiment, we used data from Uber's active riders in the Asian Pacific region between November 2019 and February 2020. 

The dataset consists of four snapshots at four different timestamps over the above period. Each snapshot has 309 features including 297 numerical features, 5 categorical features and 7 date-time features, predominantly about how each user has interacted with Uber’s different services. 304 features have missing values, and 53 of them have more than 90\% of the data points missing. Overall, the dataset has 28\% missing values. The first snapshot is used for model training and the performance of the model is measured by the rest three snapshots.

\begin{table*}[tbp]
\centering
\caption{The average test AUC scores (\%) on AutoML3 datasets}
\label{table:tab2}
\begin{tabular}{l r r r r r r r}
\toprule
 & ADA & RL & AA & B & C & D & E  \\
\midrule
Baseline & 92.14 $\pm$ 0.20 & 64.07 $\pm$ 0.41 & 70.68 $\pm$ 0.06 & 57.12 $\pm$ 0.14 & 67.47 $\pm$ 0.71 & 63.14 $\pm$ 0.15 & 81.93 $\pm$ 0.14  \\
Validation Selection & 92.12 $\pm$ 0.20 & 63.83 $\pm$ 0.47 & 68.74 $\pm$ 0.09 & 55.23 $\pm$ 0.16 & 65.08 $\pm$ 0.72 & 62.72 $\pm$ 0.12 & 81.67 $\pm$ 0.32  \\
IPW & 92.17 $\pm$ 0.22 & 62.35 $\pm$ 0.71 & 66.18 $\pm$ 0.15 & 54.47 $\pm$ 0.21 & 64.19 $\pm$ 0.75 & 60.45 $\pm$ 0.24 & 79.66 $\pm$ 0.22  \\
Feature Selection (GBDT) & 92.14 $\pm$ 0.20 & 52.44 $\pm$ 0.90 & 71.67 $\pm$ 0.18 & \textbf{60.06} $\pm$ 0.05 & 69.36 $\pm$ 0.93 & 62.11 $\pm$ 0.11 & 83.73 $\pm$ 0.03  \\
Feature Selection (DT) & 92.14 $\pm$ 0.20 & \textbf{64.82} $\pm$ 0.33 & 72.27 $\pm$ 0.04 & 59.68 $\pm$ 0.06 & 70.16 $\pm$ 0.62 & \textbf{65.61} $\pm$ 0.09 & 83.87 $\pm$ 0.04 \\
Feature Selection (RF) & 92.14 $\pm$ 0.20 & 64.41 $\pm$ 0.38 & \textbf{72.40} $\pm$ 0.04 & 59.75 $\pm$ 0.07 & \textbf{71.18} $\pm$ 0.50 & 65.26 $\pm$ 0.08 & \textbf{83.97} $\pm$ 0.02\\
\bottomrule
\end{tabular}
\end{table*}

\section{Results}
\subsection{AutoML3 for Lifelong Machine Learning Challenge Datasets}
Table~\ref{table:tab2} shows the average test AUC scores with 95\% confidence intervals of outcome classifiers across all three adversarial validation methods as well as the baseline without adversarial validation on AutoML3 datasets. The AUC scores and confidence intervals are calculated from thirty runs of experiments.

For the dataset \texttt{ADA} with the balanced training and test datasets (adversarial validation AUC $\sim$ 50\%), as expected, adversarial validation method does not affect the performance of the outcome classifier and all three methods show the same average test AUC scores as the baseline without adversarial validation.

For the remaining six AutoML3 datasets with the imbalanced training and test datasets (adversarial validation AUC $\sim$ 100\%), the feature selection method outperforms adversarial validation methods relying on propensity scores, validation data selection and IPW as well as the baseline without adversarial validation. Validation data selection and IPW perform even worse than the baseline. 

Among three training algorithms for the feature selection method, DT and RF outperform GBDT except in the dataset \texttt{B}, where GBDT performs slightly better than DT and RF. DT and RF score 1.2 $\sim$ 5.5\% (closing gap by 2.1 $\sim$ 11.4\%) better than the baseline in the average test AUC scores.

\subsection{MaLTA Datasets}
Table~\ref{table:tab3} shows training, validation, and test AUC scores with 95\% confidence intervals of outcome classifiers across all three adversarial validation methods as well as the baseline without adversarial validation on MaLTA datasets. The AUC scores and confidence intervals are calculated from thirty runs of experiments. The adversarial classifier built with Training and Test1 dataset has a perfect (100\%) AUC score, which indicates that concept drift problem exists across different time snapshots.

Among three adversarial validation methods, feature selection method with GBDT algorithm outperforms others, and enables the outcome classifier to be trained on fewer features (from 309 features to 281 features), yet achieves 3.9\% increase in average test AUC scores (closing gap by 6.3\%) over the baseline without adversarial validation. Among three training algorithms for feature selection, GBDT outperforms DT and RF unlike in AutoML3 datasets, where DT and RF outperform GBDT. 

Different results of the GBDT feature selection method between the AutoML3 and MaLTA datasets might be caused by the different proportion of missing values in the datasets. In MaLTA datasets, the majority of the features (304 out of 309 features) have missing values (overall 28\% missing), while AutoML3 datasets have only 0 $\sim$ 4\% missing values. \texttt{LightGBM}'s GBDT handles missing values differently from \texttt{scikit-learn}'s DT and RF. \texttt{scikit-learn}'s DT and RF require imputation, and we replace missing values with zero. On the other hand, \texttt{LightGBM}'s GBDT ignores missing values during splitting in each tree node and classifies them into the optimal default direction \cite{chen2016xgboost}. This way, the adversarial classifier with \texttt{LightGBM}'s GBDT might be able to detect a feature with the different distribution between the training and test datasets, even when the feature has many missing values.

Overall, Adversarial validation with feature selection improves the model performance over the baseline consistently across all seven datasets with concept drift, six in AutoML3 (except \texttt{ADA} without concept drift), and one in MaLTA.

\begin{table*}[tbp]
\centering
\caption{Training, validation and test AUC scores (\%) on MaLTA datasets}
\label{table:tab3}
\begin{tabular}{ l r r r r r }
\toprule
& Training  & Validation  & Test1  & Test2  & Test3 \\
\midrule
Baseline & 69.51 $\pm$ 0.85 & 65.27 $\pm$ 0.89 & 61.73 $\pm$ 0.47 & 62.97 $\pm$ 0.43 & 61.34 $\pm$ 0.23 \\
Validation Selection & 69.51 $\pm$ 0.71 & 65.10 $\pm$ 0.51 & 61.72 $\pm$ 0.42 & 63.04 $\pm$ 0.35 & 61.35 $\pm$ 0.19 \\
IPW & 70.97 $\pm$ 0.43 & 69.22 $\pm$ 0.08 & 62.40 $\pm$ 0.16 & 63.50 $\pm$ 0.15 & 61.51 $\pm$ 0.18 \\
Feature Selection (GBDT) & 71.34 $\pm$ 0.30 & 69.27 $\pm$ 0.04 & \textbf{64.79} $\pm$ 0.11 & \textbf{64.88} $\pm$ 0.15 & \textbf{63.05} $\pm$ 0.10\\
Feature Selection (DT) & 71.26 $\pm$ 0.28 & 69.02 $\pm$ 0.05 & 61.45 $\pm$ 0.10 & 63.75 $\pm$ 0.06  & 62.14 $\pm$ 0.05 \\
Feature Selection (RF) & 71.17 $\pm$ 0.31 & 69.00 $\pm$ 0.05 & 61.49 $\pm$ 0.51 & 63.75 $\pm$ 0.12 & 62.07 $\pm$ 0.10 \\
\bottomrule
\end{tabular}
\end{table*}

\section{Conclusion}
This paper proposes a set of new approaches that address the issue of concept drift that commonly exists in large scale user targeting automation systems, where the distribution of complex user features keeps evolving over time. The novelty of these approaches derives from the relation between concept drift and adversarial learning. Using datasets from AutoML3 Lifelong Machine Learning Challenge and Uber's user targeting automation system, MaLTA, we demonstrate that one can improve the model performance in a large scale machine learning setting by adjusting the training process through 1) adversarial feature selection, 2) validation data selection, or 3) adversarial inverse propensity weighting, all of which leverage the distributional differences between the training and test datasets. 

In addition, by comparing these different approaches, we show that adversarial feature selection consistently outperforms both validation data selection and adversarial inverse propensity weighting, while the latter two methods relying on propensity score estimation are more widely used in practice \cite{Zajac2016} and in the literature \cite{ROSENBAUM1983, Austin2011}. We conjecture that in large scale machine learning for user targeting, the high noise level in the propensity score brings more variance in the estimation, so that even though it reduces the bias, overall it increases the total risk. In contrast, adversarial feature selection, which is analogous to model selection and outlier removal in the regular machine learning, is more robust to the errors in estimating the distributional differences. In addition, the study shows the importance of choosing the right adversarial detection procedure. It suggests that one should put emphasis on extracting the discriminating features between the train and test data, instead of just minimizing the classification error, for the performance of the subsequent training task.

The study also opens new questions for further investigation. First, how does one design a validation process for optimizing the hyper-parameters to mitigate the concept drift issue? In a regular machine learning setting, one benefits from the assumption that the training set is a representative sample of the entire population. In the concept drift world, one possible solution is to assume that the concept drift follows some underlying process so that the use of backtest as a validation can be rationalized. Second, even though adversarial inverse propensity weighting shows relatively inferior performance, it will be interesting to find ways to reduce the noise of the inverse propensity weights with a combination of direct outcome regression with adversarial feature selection, similar to doubly robust estimators in the causal inference literature \cite{10.2307/2291135}. Lastly, as shown in the study, adversarial detection/estimation method plays a critical role in the overall procedure. It is a promising direction to further investigate the performance of different first stage estimation methods, or a joint estimation of adversarial detection and adversarial training in a single neural network setting (e.g. the neural network architecture in \cite{Shalit2017} for individual treatment effect estimation).

\section*{Acknowledgment}

We would like to express our appreciation to Zhenyu Zhao, Will (Youzhi) Zou and Byung-Woo Hong for their valuable and constructive feedback during the development of this work.

\bibliographystyle{ACM-Reference-Format}
\bibliography{main}

\end{document}